\documentclass[13pt]{article}
\usepackage{arxiv}
\usepackage{authblk}
\usepackage[utf8]{inputenc} 
\usepackage[T1]{fontenc}    
\usepackage{hyperref}       
\usepackage{url}            
\usepackage{booktabs}       
\usepackage{amsfonts}       
\usepackage{nicefrac}       
\usepackage{microtype}      
\usepackage{lipsum}
\usepackage{amsmath}
\usepackage{microtype}      
\usepackage{amsmath}
\usepackage{multirow}
\usepackage{amssymb}
\usepackage{graphicx}
\usepackage{wrapfig}

\newcommand{\bSigma}{\mbox{\boldmath$\Sigma$}}

\newcommand{\by}{\mbox{\boldmath$y$}}

\newcommand{\bP}{\mbox{\boldmath$P$}}
\newcommand{\bQ}{\mbox{\boldmath$Q$}}

\newcommand{\bW}{\mbox{\boldmath$W$}}
\newtheorem{lemma}{Lemma}
\newtheorem{theorem}{Theorem}[section]

\newtheorem{Algorithm}{Algorithm}[section]

\title{EPEM: Efficient Parameter Estimation for Multiple Class Monotone Missing Data}

\author[1,*]{\large Thu Nguyen}
\author[2,*]{\large Duy H. M. Nguyen}
\author[3]{\large Huy Nguyen}
\author[3, 4]{\large Binh T. Nguyen}
\author[1]{\large Bruce A. Wade}
\affil[1]{University of Louisiana at Lafayette, USA}
\affil[2]{Max Planck Institute for Informatics, Germany}
\affil[3]{University of Science, VNU-HCM, Ho Chi Minh City, Vietnam}
\affil[4]{AISIA Research Lab, Ho Chi Minh City, Vietnam}
\affil[*]{{\href{mailto:thu.nguyen@louisiana.edu}{\nolinkurl{thu.nguyen@louisiana.edu}}, 
}{\href{mailto:hnguyen@mpi-inf.mpg.de }{\nolinkurl{hnguyen@mpi-inf.mpg.de }} 
}}

\begin{document}
  \maketitle
\begin{abstract}
The problem of monotone missing data has been broadly studied during the last two decades and has many applications in different fields such as bioinformatics or statistics. Commonly used imputation techniques require multiple iterations through the data before yielding convergence. Moreover, those approaches may introduce extra noises and biases to the subsequent modeling. In this work, we derive exact formulas and propose a novel algorithm to compute the maximum likelihood estimators (MLEs) of a multiple class, monotone missing dataset when all the covariance matrices of all categories are assumed to be equal, namely EPEM. We then illustrate an application of our proposed methods in Linear Discriminant Analysis (LDA). 
As the computation is exact, our EPEM algorithm does not require multiple iterations through the data as other imputation approaches, thus promising to handle much less time-consuming than other methods. This effectiveness was validated by empirical results when EPEM reduced the error rates significantly and required a short computation time compared to several imputation-based approaches. We also release all codes and data of our experiments in one GitHub repository to contribute to the research community related to this problem.
\end{abstract}

\section{Introduction}
The problem of monotone missing data occurs in practical situations ubiquitously, especially in longitudinal studies where the information on a set of cases is collected repeatedly over time \cite{mcknight2007missing}. 
For instance, a Parkinson study \cite{hirsch2016incidence} may have multiple periods. All variables from the first block can be recorded at the beginning of the research and, therefore, fully observed.
However, the second block of variables recorded two years later, may not be fully observed as some of the patients have died. Similarly, the third block of variables, recorded three years later, can increase missing values due to more deaths. Subsequently, it can create a monotone pattern of missing data. Various other examples can be found in sensor or biological data when sensors or measurement devices, which are in data recording progress, encounter errors due to several external factors. 
Such situations usually imply a monotone missing data challenge \cite{aittokallio2010dealing, lambin2013predicting}.

Different from existing methods \cite{marivate2007autoencoder, buuren2010mice, kim2005missing, mazumder2010spectral, candes2009exact, rahman2016missing, burgette2010multiple, costa2018missing} requiring revisiting the data several times before yielding convergence and are computationally expensive in high-dimensional settings, multiple imputation replaces missing entries with substituted values, helps dealing with the problem of increased noise due to imputation, and accounts for the uncertainty in the imputations \cite{rubin1996multiple}.
In this paper, we introduce a novel algorithm, namely EPEM (\textit{Efficient Parameter Estimation for Multiple class monotone missing data}), that can compute the maximum likelihood estimators (MLEs) for both mean and covariance matrices of a monotone missing dataset when all classes' covariance matrices are assumed to be equal. Especially, our approach does not require many iterations through the data. Therefore, it is not only focused on accuracy but also has much less time-consuming than many other methods. Besides, there are potential applications in Linear Discriminant Analysis \cite{LDA}, Multiple Imputation \cite{little2019statistical}, Principle Component Analysis \cite{johnson2002applied}, and hypothesis testings \cite{little2019statistical}. It is worth noting that normal distribution is a widely used estimator when the data's distribution is unknown due to the Central Limit Theorem. Therefore, even though the data are assumed to be multivariate normal, the approach is still robust to this assumption, as illustrated in our experiments later. 

Finally, the contributions of our work can be summarized as follows: (1) We derive the exact MLEs for the mean and covariance matrix, and therefore, provide an approximation to the underlying distribution of the data; (2) We provide an efficient algorithm to compute the MLEs from the data; (3) We propose a linear discriminant procedure based on the MLEs with asymptotic properties; (4) The experimental results show that our EPEM algorithm can outperform five popular imputation methods in terms of both the error rates for parameter estimation and linear discriminant analysis classification as well as the proposed approach is superior to other techniques in terms of execution time 
when the rate of missing rate increases gradually.

The rest of this paper can be organized as follows. Section \ref{sec:related_work}  introduces  an overview of the current researches related to the problem. In Section \ref{sec:method}, we present the formulas and algorithms for computing the means and covariance matrix for the case that the data has two classes as well as for the general case of multiple categories. Section \ref{appli} is devoted to an application in linear discriminant analysis.  We illustrate the power of our approach via experiments in Section \ref{exper}. The paper ends by our conclusion and further works.

\section{Related Works}
\label{sec:related_work}
Up to now, there have been a large number of statistical and machine learning methods for missing data \cite{Bengio1995RecurrentNN,NIPS2018_7537}, which can be divided into six subcategories. The first category consists of least-square and regression-based techniques, such as, e.g., regression imputation, support vector regression \cite{marivate2007autoencoder},
multiple imputation by chained equation (MICE)\cite{buuren2010mice}, local least squares imputation \cite{kim2005missing}, and least trimmed squares imputation \cite{templ2011iterative}. 
The second category includes all techniques that are based on matrix completion such as Singular Value Decomposition (SVD) based imputation \cite{troyanskaya2001missing}, SOFT-IMPUTE \cite{mazumder2010spectral}, and Nuclear Norm Minimization \cite{candes2009exact}. 
The third category contains clustering-based techniques such as Fuzzy c-means clustering \cite{aydilek2013hybrid}, K-nearest neighbor imputation (KNN)\cite{garcia2010pattern}, fuzzy clustering-based EM imputation\cite{rahman2016missing}, imputation using fuzzy neighborhood density-based clustering \cite{razavi2016imputation}, imputation using Hybrid K-Means and Association Rules\cite{chhabra2018missing}, and evolving clustering method \cite{gautam2015data}. 
Finally, one can find techniques that are based on extreme learning, for example, e.g. AAELM \cite{gautam2015data}, extreme learning machine multiple imputation \cite{sovilj2016extreme}, or Bayesian approaches, which are Naive Bayesian imputation \cite{garcia2005naive}, Bayesian network imputation \cite{hruschka2007bayesian}, and Bayesian principal component analysis-based imputation\cite{audigier2016multiple}. 
Also, tree-based techniques such as sequential regression trees \cite{burgette2010multiple}, DMI algorithm \cite{rahman2013missing}, and C4.5 algorithm \cite{fortes2006inductive} are applied for this problem.
Deep learning techniques also form another dominant class of imputation methods. Those techniques include multiple imputations using Deep Denoising Autoencoders \cite{gondara2017multiple}, imputation via Stacked Denoising Autoencoders \cite{costa2018missing}, imputation via Adversarially-trained Graph Convolutional Networks \cite{spinelli2019missing}, a Swarm Intelligence-deep neural network \cite{leke2016missing}, combining Gravitational search algorithm with a deep-autoencoder \cite{garg2018dl}. However, a significant drawback of deep learning methods is the need for a lot of data.

There have been many results related to computations of MLEs for normal populations. Anderson \cite{anderson1957maximum} derived the MLEs for the one-group two-step monotone case. Yu et al. \cite{yu2006two} gave the MLEs for the two group three steps monotone case.
Fujisawa \cite{fujisawa1995note} derived the MLEs in closed and understandable forms for the mean vector and the covariance matrices of the general one-group monotone case by using a conditional set-up.
Kanda et al. \cite{kanda1998some} studied different properties of the MLEs for a multivariate normal population based on $k$-step monotone data. However, up to our knowledge, there has not been any study about the general case for multiple classes monotone missing data, where the data from all classes are assumed to have the same covariance matrix.  
\section{Methodology}
\label{sec:method}
In this section, we will describe our proposed approaches to calculate the maximum likelihood estimators in both two-class and multiple-class monotone missing data. 

\subsection{MLEs for Two-class Monotone Missing Data}\label{1gmle}
\subsubsection{Data Partition}
Let $\boldsymbol{D} = [\boldsymbol{y}, \boldsymbol{z}]$ be a two-class data set having missing values that can be partitioned into with the following monotone patterns:

\begin{align*} 
\boldsymbol{y}=\begin{pmatrix} \boldsymbol{y}_{11}&\dots	 &\boldsymbol{y}_{1m_{k}}&\dots&\boldsymbol{y}_{1m_2} &\dots&\boldsymbol{y}_{1m_1}\\
\boldsymbol{y}_{21} &\dots&\boldsymbol{y}_{2m_{k}}&\dots&\boldsymbol{y}_{2m_2}&\dots&*\\
\boldsymbol{y}_{31} &\dots&\boldsymbol{y}_{3m_{k}}&\dots&*&\dots&*\\
\;\;\vdots &&\;\;\vdots&\ddots&\vdots&\vdots&\vdots\\
\boldsymbol{y}_{k1}&\dots &\boldsymbol{y}_{km_{k}}&\dots&*&\dots&*
\end{pmatrix}, \\\\
\boldsymbol{z}=\begin{pmatrix} \boldsymbol{z}_{11}&\dots	 &\boldsymbol{z}_{1n_{k}}&\dots&\boldsymbol{z}_{1n_2} &\dots&\boldsymbol{z}_{1n_1}\\
\boldsymbol{z}_{21} &\dots&\boldsymbol{z}_{2n_{k}}&\dots&\boldsymbol{z}_{2n_2}&\dots&*\\
\boldsymbol{z}_{31} &\dots&\boldsymbol{z}_{3n_{k}}&\dots&*&\dots&*\\
\;\;\vdots &&\;\;\vdots&\ddots&\vdots&\vdots&\vdots\\
\boldsymbol{z}_{k1}&\dots &\boldsymbol{z}_{kn_{k}}&\dots&*&\dots&*
\end{pmatrix},
\end{align*}

Here, for all $i=1,\dots,k$, $p_i$ is the number of features of a sample $\boldsymbol{y}_{i1}$ (note that $\boldsymbol{y}_{i1}, \boldsymbol{y}_{i2},\dots, \boldsymbol{y}_{in_i}$ have the same number of features), $n_i$ is the number of samples of $\boldsymbol{y}_{i1}$, and $k$ is the number of data blocks. We assume that all samples in $\boldsymbol{y}$ belong to the first class, all samples in $\boldsymbol{y}$ belong to the second class, and each ``$*$'' denotes a missing block of values. It means there are $n_1, m_1$ observations available on the first $p_1$ variables, $n_2, m_2$ observations available on the  first $p_1+p_2$ variables, and so on, for the first and the second classes, respectively. Typically, one can partition $\boldsymbol{y}$ into:

\begin{align*} 
	\boldsymbol{y}_{1}&=&\begin{pmatrix}
		\boldsymbol{y}_{11}& \dots &\boldsymbol{y}_{1n_k} & \dots &\boldsymbol{y}_{1n_2}& \dots & \boldsymbol{y}_{1n_1}
	\end{pmatrix}_{p_1\times n_1 },
	\\
	\boldsymbol{y}_{2} &=&\begin{pmatrix}
		\boldsymbol{y}_{11}  & \dots	& \boldsymbol{y}_{1n_k}&\dots	& \boldsymbol{y}_{1n_2}\\
		\boldsymbol{y}_{21}&\dots & \boldsymbol{y}_{2n_k} &\dots& \boldsymbol{y}_{2n_2}
	\end{pmatrix}_{(p_1+p_2)\times n_2},
	\\
	\vdots&&\\
	\boldsymbol{y}_{k} &=&\begin{pmatrix}
		\boldsymbol{y}_{11} &\dots&  \boldsymbol{y}_{1n_k}\\
		\boldsymbol{y}_{21}&\dots& \boldsymbol{y}_{2n_k}\\
		\vdots & \ddots & \vdots \\
		\boldsymbol{y}_{k1}&\dots&\boldsymbol{y}_{kn_k}
	\end{pmatrix}_{(\sum_{j=1}^{k}p_j)\times n_k}.
\end{align*} 
Similarly, one can determine $k$ partitions of $\boldsymbol{z}$, including $\boldsymbol{z}_{1}$, $\boldsymbol{z}_{2}$, $\dots$, $\boldsymbol{z}_{k}$. 

\subsubsection{Maximum Likelihood Estimators}
For the rest of the paper, we denote $\hat{\theta}$ as the maximum likelihood estimate of a parameter $\theta$.
Suppose that the data from the first class ($\boldsymbol{y}$) follow a multivariate normal distribution with mean $\boldsymbol {\mu }$, and the data from the second class ($\boldsymbol{z}$) follow a multivariate normal distribution with mean $\boldsymbol{\eta}$, and both classes have the same covariance matrix $\mathbf{\Sigma}$.  
Let $\bar{\boldsymbol{y}}_{i}$ and $\mathbf{S}_i$ be the mean and  covariance matrices of $\boldsymbol{y}_{i}$,  $\bar{\boldsymbol{z}}_{i}$ and $\mathbf{R}_i$ be the mean and covariance matrices of $\boldsymbol{z}_{i}$, respectively. Then, one can define

\begin{equation}
\mathbf{W}_i =  (n_i-1)\mathbf{S}_i+ (m_i-1)\mathbf{R}_i,
\end{equation}

as the sum of square and cross product matrix of $\boldsymbol{y}_i$ and  $\mathbf{z}_i$, and $\boldsymbol\Sigma_i$ as the leading principal submatrix of $\boldsymbol\Sigma$ of size $\left(\sum_{j=1}^{i}p_j\right) \times \left(\sum_{j=1}^{i}p_j\right)$, for any $i \in \{1,2,\dots,k\}$. We denote 

\begin{equation}
\boldsymbol{\mu} = \begin{pmatrix}
[\boldsymbol{\mu} ]_1\\
[\boldsymbol{\mu}]_2\\
\vdots\\
[\boldsymbol{\mu}]_k
\end{pmatrix},
\boldsymbol{\mu}_i = \begin{pmatrix}
[\boldsymbol{\mu} ]_1\\
[\boldsymbol{\mu}]_2\\
\vdots\\
[\boldsymbol{\mu}]_i
\end{pmatrix},
\end{equation}

where $[\boldsymbol{\mu}]_j$ is a vector of order $p_j \times 1$ ($j=1,\dots,k$).
To represent $\widehat{\boldsymbol {\Sigma }}_i$ in terms of $\widehat{\boldsymbol {\Sigma }}_{i-1}$ as well as $\widehat{\boldsymbol {\mu }}_i$ in terms of $\widehat{\boldsymbol {\mu }}_{i-1}$, we partition 
$\bar{\boldsymbol{x}}_{i}$,$\boldsymbol \Sigma_i$, and $\mathbf{W}_i$ as follows:

\begin{equation}
\bar{\boldsymbol{y}}_{i} = \begin{pmatrix}
{[\bar{\boldsymbol{y}}_{i}]_1}\\
{[\bar{\boldsymbol{y}}_{i}]_2}
\end{pmatrix}, {\rm } 
\boldsymbol\Sigma_i=\begin{pmatrix}
[\boldsymbol\Sigma_i]_{11} & [\boldsymbol\Sigma_i]_{12}\\
[\boldsymbol\Sigma_i]_{21} & [\boldsymbol\Sigma_i]_{22}
\end{pmatrix},\\
\mathbf{W}_i=\begin{pmatrix}
[\mathbf{W}_i ]_{11} & [\mathbf{W}_i]_{12}\\
[\mathbf{W}_i]_{21} & [\mathbf{W}_i]_{22}
\end{pmatrix} \ \ i=2,\dots,k,
\end{equation}

where $[\bar{\boldsymbol{y}}_{i}]_1$ and ${[\bar{\boldsymbol{y}}_{i}]_2}$ have the size of  ${(\sum_{j=1}^{i-1}p_j)\times 1}$,  ${p_i\times 1}$, respectively;  $[\boldsymbol\Sigma_i]_{11}$ and $[\mathbf{W}_i]_{11}$ have the size $\left(\sum_{j=1}^{i-1}p_j\right)\times \left(\sum_{j=1}^{i-1}p_j\right)$; $[\boldsymbol\Sigma_i]_{12}$ and $[\mathbf{W}_i]_{12}$ are matrices of size $\left(\sum_{j=1}^{i-1}p_j\right)\times p_i$;  $[\boldsymbol\Sigma_i]_{22}$ and $[\mathbf{W}_i]_{22}$ have the size $p_i \times p_i$.
Next, we can create similar partitions for  $\bar{\boldsymbol{z}}_i$ and $ \boldsymbol{\eta}$, and obtain the following theorem:

\begin{theorem}
	\label{2cl}
	With given notations, the MLEs for the two-class case can be explicitly expressed as:
	\begin{align*}
	\label{mle-mu}
	\widehat{\boldsymbol{\mu}}_1 =\bar{\boldsymbol{y}}_1,\;\;\;
	[\widehat{\boldsymbol{\mu}}]_i = [\bar{\boldsymbol{y}}_i]_2 - \bP_i([\bar{\boldsymbol{y}}_i]_1 - \widehat{\boldsymbol{\mu}}_{i-1}),\\
	\widehat{\boldsymbol{\eta}}_1  =\bar{\boldsymbol{z}}_1,\;\;\;
	[\widehat{\boldsymbol{\eta}}]_i = [\bar{\boldsymbol{z}}_i]_2 - \bP_i([\bar{\boldsymbol{z}}_i]_1 - \widehat{\boldsymbol{\eta}}_{i-1}),
	\end{align*}
	and for $i=2,.., k$,
	\begin{align}
	\widehat{\boldsymbol\Sigma}_1&=\frac{{\mathbf{W}}_1}{n_1+m_1}, 
	\widehat{[\boldsymbol\Sigma_i]}_{21} = \bP_i\widehat{[\boldsymbol\Sigma_i]}_{11},\\
	\widehat{[\boldsymbol\Sigma_i]}_{12}&=\widehat{[\boldsymbol\Sigma_i]}_{21}',
	\widehat{[\boldsymbol\Sigma_i]}_{22} = \bQ_i + \bP_i \widehat{[\boldsymbol\Sigma_i]}_{12},
	\end{align} 
	where  
	\begin{equation}\label{sum-stats-1}
	\boldsymbol P_i = [\mathbf{W}_i]_{21}[\mathbf{W}_i]_{11}^{-1}\;\;\;\text{and}\;\;\;
	\bQ_i = \frac{1}{ n_i+m_i}\left([\mathbf{W}_i]_{22} -\bP_i [\mathbf{W}_i]_{12} \right).
	\end{equation}
\end{theorem}

All details of the proof of Theorem \ref{2cl} can be found in the Appendix \ref{sec:appendix_B}.
Based on Theorem \ref{2cl}, we construct the following algorithm, namely EPEM (\textit{Efficient Parameter Estimation for Multiple class monotone missing data}), to calculate the MLEs for the two-class monotone missing data.
\begin{Algorithm}{(EPEM algorithm for two-class monotone missing data)} 
\begin{enumerate}
		\item Input: $\boldsymbol{y}, \mathbf{z}, k; n_i, m_i, p_i$, for $i=1,2,..,k$.
		
		\item Initiate: $\widehat{\boldsymbol{\mu}}\leftarrow \bar{\boldsymbol{y}}_1, \widehat{\boldsymbol{\eta}}\leftarrow \bar{\boldsymbol{z}}_1,
		\widehat{\bSigma}\leftarrow {\mathbf{W}_1}/\left({n_1+m_1}\right)$.
		
		\item  For $2\le i\le k$:
		\begin{itemize}
			\item Compute: 
\begin{equation}
	\bP_i \leftarrow [\mathbf{W}_i]_{21}[\mathbf{W}_i]_{11}^{-1} \text{ and }
	\bQ_i \leftarrow  \frac{1}{ n_i+m_i}\left([\mathbf{W}_i]_{22} -\bP_i [\mathbf{W}_i]_{12}\right).
\end{equation}			
			\item Update: 
			\begin{eqnarray}
			\widehat{[\bSigma_i]}_{21} &\leftarrow&  \bP_i\widehat{[\bSigma_i]}_{11}, \boldsymbol{\widehat{\mu}} \leftarrow  \begin{pmatrix}
			\widehat{\boldsymbol{\mu}}\\ [\bar{\boldsymbol{y}}_i]_2 - \bP_i([\bar{\boldsymbol{y}}_i]_1 -  \widehat{\boldsymbol{\mu}})
			\end{pmatrix}  \nonumber\\
			\widehat{\boldsymbol{\eta}}&\leftarrow&  \begin{pmatrix}
			\widehat{\boldsymbol{\eta}}\\ [\bar{\boldsymbol{z}}_i]_2 - \bP_i([\bar{\boldsymbol{y}}_i]_1 - \widehat{\boldsymbol{\eta}})
			\end{pmatrix},
			\widehat{\boldsymbol {\Sigma }}  \leftarrow  \begin{pmatrix}
			\widehat{\bSigma} &\widehat{[\bSigma_i]}_{21}'\\
			\widehat{[\bSigma_i]}_{21} & \bQ_i+\bP_i\widehat{[\bSigma_i]}_{21}'  \nonumber
\end{pmatrix}
			\end{eqnarray}
		\end{itemize}
		\item Output: $\widehat{\boldsymbol {\mu }}, \widehat{\boldsymbol {\eta} }, \widehat{\boldsymbol {\Sigma }}$.
\end{enumerate}
\end{Algorithm}

\subsection{MLEs for Multiple-class Monotone Missing Data}
\label{mgmle}
Assume that there is a dataset with $G$ categories (classes), where each sample from class $g^{th}( 1\le g\le G)$ follows a $p-$ dimensional multivariate normal distribution with mean $\boldsymbol{\mu}^{(g)}$ and covariance $\bSigma$, denoted as $N_p(\boldsymbol{\mu}^{(g)},\bSigma)$. In addition, let each sample $\boldsymbol{x}^{(g)}$ from class $g^{th}$ have the following monotone pattern:
\begin{align*}
\boldsymbol{x}^{(g)}=\begin{pmatrix} \boldsymbol{x}_{11}^{(g)}&\dots	 &\boldsymbol{x}_{1n_{k}^g}^{(g)}&\dots	 &\boldsymbol{x}_{1n_3^g}^{(g)}&\boldsymbol{x}_{1n_2^g}^{(g)} &\boldsymbol{x}_{1n_1^g}^{(g)}\\
\boldsymbol{x}_{21}^{(g)} &\dots&\boldsymbol{x}_{2n_{k}^g}^{(g)}&\dots&\boldsymbol{x}_{2n_3^g}^{(g)}&\boldsymbol{x}_{2n_2^g}^{(g)}&*\\
\boldsymbol{x}_{31}^{(g)} &\dots&\boldsymbol{x}_{3n_{k}^g}^{(g)}&\dots&\boldsymbol{x}_{3n_2^g}^{(g)}&*&*\\
\vdots && \vdots&\ddots&\vdots&\vdots&\vdots\\
\boldsymbol{x}_{k1}^{(g)}&\dots &\boldsymbol{x}_{kn_{k}^g}^{(g)}&\dots&*&*&*
\end{pmatrix}.
\end{align*}

That is, there are $n_1^{(g)}$ observations available on the first $p_1^{(g)}$ variables, $n_2^{(g)}$ observations available on the  first $p_1^{(g)}+p_2^{(g)}$ variables, and so on. For $1\le g\le G$, one can partition the data in group $g^{th}$ into:

\begin{eqnarray}
\boldsymbol{x}^{(g)}_{1}&=&\begin{pmatrix}
\boldsymbol{x}_{11}^{(g)}& \dots &\boldsymbol{x}_{1n_k^{(g)}} ^{(g)}& \hdots &\boldsymbol{x}_{1n_2^{(g)}}^{(g)}& \dots & \boldsymbol{x}_{1n_1^{(g)}}^{(g)}
\end{pmatrix}, \nonumber \\
\boldsymbol{x}^{(g)}_{2} &=&\begin{pmatrix}
\boldsymbol{x}_{11}^{(g)}  & \dots	& \boldsymbol{x}_{1n_k^{(g)}}^{(g)}&\dots	& \boldsymbol{x}_{1n_2^{(g)}}^{(g)}\\
\boldsymbol{x}_{21}^{(g)}&\dots & \boldsymbol{x}_{2n_k^{(g)}}^{(g)} &\dots& \boldsymbol{x}_{2n_2^{(g)}}^{(g)}
\end{pmatrix},  \nonumber \\
\vdots  \nonumber \\
\boldsymbol{x}^{(g)}_{k}&=&\begin{pmatrix}
\boldsymbol{x}_{11}^{(g)} &\dots&  \boldsymbol{x}_{1n_k^{(g)}}^{(g)}  \nonumber \\
\boldsymbol{x}_{21}^{(g)}&\dots& \boldsymbol{x}_{2n_k^{(g)}}^{(g)}  \nonumber \\
\vdots & \ddots & \vdots \\
\boldsymbol{x}_{k1}^{(g)}&\dots&\boldsymbol{x}_{kn_k^{(g)}}^{(g)}  \nonumber
\end{pmatrix},
\end{eqnarray}

where $\boldsymbol{x}^{(g)}_{1}$ has the size ${p_1^{(g)}\times n_1^{(g)} }$, $	\boldsymbol{x}^{(g)}_{2}$ is of size ${(p_1^{(g)}+p_2^{(g)})\times n_2^{(g)}}$, \dots, $\boldsymbol{x}^{(g)}_{k}$ has the size ${(\sum_{j=1}^{k}p_j^{(g)})\times n_k^{(g)}}$. Here, we assume that $p_i^{(g)} = p_i$ for all $i=1,\dots,k$ and $g=1,\dots,G$.

Let $\bar{\boldsymbol{x}}^{(g)}_{i}, \mathbf{S}^{(g)}_i$ be the mean and covariance matrix of $\boldsymbol{x}^{(g)}_{i}$, respectively.
For $i=1,\dots,k$, we define 
\begin{align}
\bW_i &= \sum_{g=1}^{G}(n_i^{(g)}-1) \mathbf{S}_i^{(g)},
\boldsymbol {\mu }=(\boldsymbol {\mu }^{(1)},\dots,\boldsymbol {\mu }^{(G)}),\;\;\;\bar{\boldsymbol{x} }_i = (\bar{\boldsymbol{x} }^{(1)},\dots,\bar{\boldsymbol{x} }^{(G)})
\end{align}
as the sum of squares and cross products matrix, the matrix of all classes' means, and the matrix of all classes' sample means. Now, we regard $\bSigma_i$ as the $(\sum_{j=1}^{i}p_j)^{th}$ order leading principal submatrix of $\bSigma$, for $i=1,\dots,k$, and do the following partitions:
\begin{equation}
\bar{\boldsymbol{x}}_{i} = \begin{pmatrix}
{[\bar{\boldsymbol{x}}_{i}]_1}\\
{[\bar{\boldsymbol{x}}_{i}]_2}
\end{pmatrix},\;\;\;
\boldsymbol{\mu} =  \begin{pmatrix}
[\boldsymbol{\mu}]_1\\
[\boldsymbol{\mu}]_2\\
\vdots\\
[\boldsymbol{\mu}]_k
\end{pmatrix},  
\end{equation}
We denote
\begin{equation}
\bSigma_i=\begin{pmatrix}
[\bSigma_i]_{11} & [\bSigma_i]_{12}\\
[\bSigma_i]_{21} & [\bSigma_i]_{22}
\end{pmatrix},\;\;\;
\bW_i=\begin{pmatrix}
[\bW_i]_{11} & [\bW_i]_{12}\\
[\bW_i]_{21} & [\bW_i]_{22}
\end{pmatrix}, \ \ (i=2,\dots,k),
\end{equation}
where $[\bar{\boldsymbol{x}}_{i}]_1$ contains the first  ${\sum_{j=1}^{i-1}p_j}$ rows of $\bar{\boldsymbol{x}}_{i}$, and ${[\bar{\boldsymbol{x}}_{i}]_2}$ has all ${p_i}$ remaining rows; $[\boldsymbol{\mu}^{(g)}]_j$ is a block of order $p_j \times \left(\sum_{j=1}^kp_j\right)$, $j=1,\dots,k$; 
$[\bSigma_i]_{11}$ and $[\bW_i]_{11}$ are of order $\left(\sum_{j=1}^{i-1}p_j\right)\times \left(\sum_{j=1}^{i-1}p_j\right)$, $[\bSigma_i]_{12}$ and $[\bW_i]_{12}$ are of order $\left(\sum_{j=1}^{i-1}p_j\right)\times p_i$, $[\bSigma_i]_{22}$ and $[\bW_i]_{22}$ are of order $p_i \times p_i$.

Now, if 
\begin{equation}
\boldsymbol {\mu }_i=\begin{pmatrix}
[\boldsymbol {\mu } ]_1\\
\vdots\\
[\boldsymbol {\mu } ]_i.
\end{pmatrix}, i=1,\dots,k,
\end{equation}
then we have the following theorem:
\begin{theorem}\label{general}
	With given notations, the MLEs are calculated by the following recurrent equations:
	\begin{equation}
	\widehat{\boldsymbol{\mu}}_1  = \bar{\boldsymbol{x}}_1,      \;\;\;   
	[\widehat{\boldsymbol{\mu}}]_i= [\bar{\boldsymbol{x}}_i]_2 - \bP_i([\bar{\boldsymbol{x}}_i]_1 - \widehat{\boldsymbol{\mu}}_{i-1}), 
	\widehat{\bSigma}_1=\frac{\bW_1}{\sum_{g=1}^G n_1^g},  	\widehat{[\bSigma_i]}_{21} = \bP_i\widehat{[\bSigma_i]}_{11}, 
	\end{equation}
	\begin{eqnarray}
	\label{pooled_mles}	     
	\widehat{[\bSigma_i]}_{12}&=(\widehat{[\bSigma_i]}_{21})',  
	\widehat{[\bSigma_i]}_{22} &= \bQ_i + \bP_i \widehat{[\bSigma_i]}_{12},
	\end{eqnarray}
	for $i=2,\dots, k$, where
	\begin{equation}
	\bP_i =   [\bW_i]_{21}  [\bW_i]_{11} ^{-1}\;\;\;\text{and}\;\;\;
	\bQ_i =\frac{1}{\sum_{g=1}^G n_i^{(g)}}\left([\bW_i]_{22} -\bP_i [\bW_i]_{12}\right).
	\end{equation}
\end{theorem}

We provide the proof of Theorem \ref{general} in the Appendix \ref{sec:appendix_C}. From this theorem, we design the following algorithm to compute MLEs for multiple-class monotone missing data:

\begin{Algorithm}{(EPEM algorithm for multiple -class monotone missing data)} 
	\begin{enumerate}
		\item Input: $\boldsymbol{x}^{(g)}, n_i^{(g)}, p_i$, for $g=1,2,.., G; \;i = 1,2,\dots, k$.
		
		\item Initiate: $\widehat{\boldsymbol{\mu}}\leftarrow \bar{\boldsymbol{x}}_1, \widehat{\bSigma}\leftarrow {\bW_1}/\left({\sum_{g=1}^Gn_1^{(g)}}\right)$.
		
		\item  For $2\le i\le k$:
		\begin{itemize}
			\item Compute: $\bP_i = [\bW_i]_{21}[\bW_i]_{11}^{-1}$, $\bQ_i \leftarrow  \frac{1}{ \sum_{g=1}^Gn_i^{(g)}}\left([\bW_i]_{22}-\bP_i [\bW_i]_{12} \right)$.
			
			\item Update: $\hat{\boldsymbol {\mu } }\leftarrow  \begin{pmatrix}\hat{\boldsymbol {\mu } }\\ [\bar{\boldsymbol{x} }_i]_2 - P_i([\bar{\boldsymbol{x} }_i]_1-\hat{\boldsymbol {\mu } })
			\end{pmatrix}$,
			$\widehat{\bSigma} \leftarrow  \begin{pmatrix}
			\widehat{\bSigma} & \widehat{[\bSigma_i]}_{21}' \nonumber\\
			\widehat{[\bSigma_i]}_{21} & \bQ_i+\bP_i\widehat{[\bSigma_i]}_{21}'. \nonumber
			\end{pmatrix}$
		\end{itemize}
		\item Output: $\widehat{\boldsymbol {\mu } }, \widehat{\boldsymbol {\Sigma }}$.
	\end{enumerate}
\end{Algorithm}

Importantly, we can ensure that:
\begin{theorem}
	\label{theo:3}
	The resulting estimates of the EPEM algorithm are asymptotic, i.e.,
	\begin{equation}
	\hat{\boldsymbol {\mu } }\overset{p}{\rightarrow} \boldsymbol {\mu }, \text{and} \;\;\;	\hat{\boldsymbol {\Sigma } }\overset{p}{\rightarrow} \boldsymbol {\Sigma }, 
	\end{equation}
	where $\overset{p}{\rightarrow}$ denotes convergence in probability.
\end{theorem}
One can see our proof for Theorem \ref{theo:3} in the Appendix \ref{sec:appendix_D}.

\section{Applications in Linear Discriminant Analysis with Monotone Missing Data} \label{appli}
Suppose that there are given data of $G$ classes, and data from each class follows a multivariate normal distribution with mean $\boldsymbol {\mu } _k$ and covariance $\boldsymbol {\Sigma }$, i.e., the data from all classes have the same covariance matrix. In LDA, the classification rule that minimizes the total probability of missclassification is to assign $\boldsymbol{x}$ to $\pi_h$ if
\begin{equation}\label{lda}
{d}_h(\boldsymbol{x}) = \max\{{d}_1(\boldsymbol{x}),{d}_2(\boldsymbol{x}),\dots,{d}_G(\boldsymbol{x}) \},
\end{equation}
where
\begin{equation*}
{d}_i(\boldsymbol{x}) = \boldsymbol{\mu}_i'\bSigma^{-1}\boldsymbol{x}-\frac{1}{2}\boldsymbol{\mu}_i'\bSigma^{-1}\boldsymbol{\mu}_i+\ln r_i, \;\; i=1,2,\dots,G.
\end{equation*}
Here, $r_i$ is the proportion of the number of samples that belong to the $i^{th}$ class.
In our approach, the parameters $\boldsymbol {\mu } _i, \boldsymbol {\Sigma } $ are replaced by the corresponding maximum likelihood estimates. Therefore, the classification rule is to assign $\boldsymbol{x}$ to class $\pi_h$ if
\begin{equation}\label{lda1}
\hat{d}_h(\boldsymbol{x}) = \max\{\hat{d}_1(\boldsymbol{x}),\hat{d}_2(\boldsymbol{x}),\dots,\hat{d}_G(\boldsymbol{x}) \}, 
\end{equation}
where for $i=1,2,\dots,G$,
\begin{equation}\label{dix}
\hat{d}_i(\boldsymbol{x}) = \hat{\boldsymbol{\mu}}_i'\widehat{\bSigma}^{-1}\boldsymbol{x}-\frac{1}{2}\hat{\boldsymbol{\mu}}_i'\widehat{\bSigma}^{-1}\hat{\boldsymbol{\mu}}_i+\ln \hat{r}_i.
\end{equation}
We have the following important results:
\begin{theorem}\label{mleinv}
	\textbf{(Invariant under Non-degenerate Linear Transformations)}
	The classification rule (\ref{lda1}) is invariant to non-degenerate linear transformations, i.e., if we transform the data by an invertible linear transformation $A$ then the classification rule is the same as before transforming, i.e., to assign $\boldsymbol{x}$ to $\pi_h$ if
	\begin{align}
	\hat{d}_h(\boldsymbol{x}) = \max\{\hat{d}_1(\boldsymbol{x}),\hat{d}_2(\boldsymbol{x}),\dots,\hat{d}_G(\boldsymbol{x}) \}
	\end{align}
	where $\hat{d}_i(\boldsymbol{x})$ is defined as in Eq. (\ref{dix}) for $i=1,2,\dots, G$.
\end{theorem}

\begin{theorem}\label{mleasym}
	$\hat{d}_i\left(\boldsymbol{x}\right)$ converges in probability to $d_i\left(\boldsymbol{x}\right)$ for all $i=1,2,\ldots,G$.
\end{theorem}
All details of the proof for both Theorems \ref{mleinv} and \ref{mleasym} can be provided in Appendix \ref{sec:appendix_E} and Appendix \ref{sec:appendix_F}.

\section{Experiments}\label{exper}
In this section, we describe all datasets, the implementation of the proposed techniques, and experimental results in detail. 
\subsection{ Settings}
To illustrate the efficiency of our algorithm, we compare the results of the proposed method with other imputation methods, including imputation based on Expectation Maximization (EM)\cite{dempster1977maximum}, K-nearest neighbor (KNN) imputation \cite{garcia2010pattern}, Multiple Imputation by Chained Equation (MICE) \cite{buuren2010mice}, SOFT-IMPUTE \cite{mazumder2010spectral}, and imputation by Nuclear Norm Minimization \cite{candes2009exact}.

\begin{table*}[ht]
  \centering
  	\label{tab1}
  	\scalebox{0.9}{
  \begin{tabular}{cccc}
    \toprule
    \cmidrule(r){1-4}
    \textbf{Dataset}     & \textbf{Classes} &  \textbf{ Features} & \textbf{Samples} \\
    \midrule
    Iris	 & 3	 &  4     & 150\\
    Wine	 & 3	 &  13        & 178\\
    Seeds & 3 & 7 & 210 \\
    Parkinson & 2 & 22 & 195\\
    Digits & 10 & 64 & 1797\\
    Inosphere & 2 & 34 & 351\\
    \bottomrule
  \end{tabular}}
   \caption{The description of UCI data sets that are used in our experiments.}
\end{table*}

We perform the experiments on five distinct data sets from the Machine Learning Database Repository at the University of California, Irvine \cite{Dua:2019}: Iris, Wine, Seeds, Parkinson and Digits. Table \ref{tab1} shows a summary of those data sets. In each dataset, we normalize by removing the mean and scaling to unit variance all features. Furthermore, in the Digits dataset that contains $1797$ images with size $8\times 8$, we 
eliminate ten columns whose entries are 0 almost everywhere. For the Inosphere data set, we delete the first column where the values are the same within one of the classes, and the second columns where all entries are 0. 
To evaluate of the LDA classification task, we propose the following metric to measure the performance of the parameter estimation process:
\begin{equation}
r = \frac{||\boldsymbol {\mu } -\boldsymbol {\hat{\mu}}||_F }{n_{\boldsymbol {\mu } }}
+\frac{||\boldsymbol {\Sigma
	} -\boldsymbol {\hat{\Sigma}}||_F }{n_{\boldsymbol {\Sigma } }},
\end{equation}
where $||.||$ denotes the Frobenius norm; $\hat{\boldsymbol {\mu } }, \hat{\boldsymbol {\Sigma } }$ are estimated values derived from EPEM for $\boldsymbol {\mu } , \boldsymbol {\Sigma } $ respectively; ${n_{\boldsymbol {\mu } }}$,   ${n_{\boldsymbol {\Sigma } }}$ are the corresponding number of entries in $\boldsymbol {\mu }, \boldsymbol {\Sigma }$ and the ground truth $\boldsymbol {\mu }, \boldsymbol {\Sigma }$ are calculated from the full data without missing values. For interpretation, note that this is the sum of the average difference of each entry in the mean/common covariance matrix. 

\begin{table*}[ht]
	\vspace{0.1in}
	\label{paraerr}
	\centering
	\small
	\scalebox{0.9}{
		\begin{tabular}{|c|c|c|c|c|c|c|c|}
	    \toprule
			\textbf{Dataset}                    & \textbf{\begin{tabular}[c]{@{}c@{}}Missing\\ Rate(\%)\end{tabular}} & \textbf{EPEM} & \textbf{\begin{tabular}[c]{@{}c@{}}KNN\\ Imputation\end{tabular}} & \textbf{MICE} &
			\textbf{\begin{tabular}[c]{@{}c@{}}SOFT-\\ IMPUTE\end{tabular}} & \textbf{\begin{tabular}[c]{@{}c@{}}EM\\ Imputation\end{tabular}} & \textbf{\begin{tabular}[c]{@{}c@{}}Nuclear  \\ Norm\end{tabular}} \\ \hline
			    \midrule
			\multirow{3}{*}{Seeds}     & 20\%                                                                & \textbf{0.016} & 0.046                                                             & 0.019                                                              & 0.024                                                           & 0.046                                                            & 0.024                                                                         \\ 
			& 30\%                                                                & \textbf{0.020} & 0.065                                                             & 0.025                                                              & 0.032                                                           & 0.070                                                            & 0.031                                                                         \\ 
			& 40\%                                                                & \textbf{0.023} & 0.081                                                             & 0030                                                               & 0.045                                                           & 0.078                                                            & 0.040                                                                         \\ \hline
			\multirow{3}{*}{Iris}      & 20\%                                                                & 0.027          & 0.066                                                             & \textbf{0.025}                                                     & 0.038                                                           & 0.069                                                            & 0.037                                                                         \\ 
			& 30\%                                                                & \textbf{0.031} & 0.091                                                             & 0.039                                                              & 0.059                                                           & 0.105                                                            & 0.058                                                                         \\ 
			& 40\%                                                                & \textbf{0.033} & 0.123                                                             & 0052                                                               & 0.088                                                           & 0.140                                                            & 0.084                                                                         \\ \hline
			\multirow{3}{*}{Parkinson} & 20\%                                                                & \textbf{0.025} & 0.087                                                             & 0.095                                                              & 0.090                                                           & 0.088                                                            & 0.091                                                                         \\ 
			& 30\%                                                                & \textbf{0.026} & 0.087                                                             & 0.091                                                              & 0.090                                                           & 0.088                                                            & 0.090                                                                         \\ 
			& 40\%                                                                & \textbf{0.043} & 0.101                                                             & 0.119                                                              & 0.102                                                           & 0.099                                                            & 0.102                                                                         \\ \hline
			\multirow{3}{*}{Wine}      & 20\%                                                                & \textbf{0.018} & 0.034                                                             & 0.024                                                              & 0.030                                                           & 0.035                                                            & 0.029                                                                         \\ 
			& 30\%                                                                & \textbf{0.024} & 0.042                                                             & 0.030                                                              & 0.035                                                           & 0.049                                                            & 0.035                                                                         \\ 
			& 40\%                                                                & \textbf{0.031} & 0.057                                                             & 0.045                                                              & 0.051                                                           & 0.063                                                            & 0.050                                                                         \\ \hline
			\multirow{3}{*}{Digits}    & 20\%                                                                & \textbf{0.003} & 0.026                                                             & 0.026                                                              & 0.026                                                           & 0.026                                                            & 0.026                                                                         \\ 
			& 30\%                                                                & \textbf{0.009} & 0.028                                                             & 0.028                                                              & 0.028                                                           & 0.028                                                            & 0.028                                                                         \\ 
			& 40\%                                                                & \textbf{0.009} & 0.027                                                             & 0.028                                                              & 0.027                                                           & 0.028                                                            & 0.027                                                                         \\ \hline
			\multirow{3}{*}{Inosphere} & 20\% &\textbf{0.011} & 0.029 & 0.029 & 0.029 & 0.030 & 0.029 \\
			& 30\% &\textbf{ 0.011} & 0.030 & 0.029 & 0.029 & 0.030 & 0.029\\
			& 40\% & \textbf{0.013} & 0.031 & 0.029 & 0.030 & 0.032 & 0.030\\\hline
	\end{tabular}}
		\caption{Parameters estimation errors with different missing rates}
\end{table*}

\begin{table*}[ht]
	\vspace{0.1in}
	\label{cverr}
	\centering
	\small	
	\scalebox{0.9}{
		\begin{tabular}{|c|c|c|c|c|c|c|c|}
			\toprule
			\textbf{Dataset}                    & \textbf{\begin{tabular}[c]{@{}c@{}}Missing\\ Rate(\%)\end{tabular}} & \textbf{EPEM} & 
			\textbf{\begin{tabular}[c]{@{}c@{}}KNN\\ imputation\end{tabular}} & \textbf{MICE}& \textbf{\begin{tabular}[c]{@{}c@{}}SOFT-\\ IMPUTE\end{tabular}} & \textbf{\begin{tabular}[c]{@{}c@{}}EM\\ imputation\end{tabular}} & \textbf{\begin{tabular}[c]{@{}c@{}}Nuclear\\ Norm\end{tabular}} \\ \hline
			\midrule
			\multirow{3}{*}{Seeds}     & 20\%                                                                & \textbf{0.034} & 0.051                                                             & 0.042                                                              & 0.052                                                           & 0.086                                                            & 0.052                                                           \\ 
			& 30\%                                                                & \textbf{0.038} & 0.051                                                             & 0.062                                                              & 0.062                                                           & 0.086                                                            & 0.062                                                           \\ 
			& 40\%                                                                & \textbf{0.038} & 0.071                                                             & 0.078                                                              & 0.074                                                           & 0.094                                                            & 0.068                                                           \\ \hline
			\multirow{3}{*}{Iris}      & 20\%                                                                & \textbf{0.024} & 0.125                                                             & 0.063                                                              & 0.117                                                           & 0.166                                                            & 0.117                                                           \\ 
			& 30\%                                                                & \textbf{0.032} & 0.155                                                             & 0.063                                                              & 0.126                                                           & 0.153                                                            & 0.135                                                           \\ 
			& 40\%                                                                & \textbf{0.037} & 0.153                                                             & 0.075                                                              & 0.133                                                           & 0.164                                                            & 0.133                                                           \\ \hline
			\multirow{3}{*}{Parkinson} & 20\%                                                                & \textbf{0.146} & 0.160                                                             & 0.182                                                              & 0.178                                                           & 0.182                                                            & 0.187                                                           \\ 
			& 30\%                                                                & \textbf{0.152} & 0.187                                                             & 0.190                                                              & 0.182                                                           & 0.168                                                            & 0.201                                                           \\ 
			& 40\%                                                                & 0.187          & 0.211                                                             & 0.199                                                              & 0.178                                                           & 0.178                                                            & \textbf{0.173}                                                  \\ \hline
			\multirow{3}{*}{Wine}      & 20\%                                                                & \textbf{0.011} & 0.025                                                             & 0.018                                                              & 0.013                                                           & 0.030                                                            & 0.013                                                           \\ 
			& 30\%                                                                & \textbf{0.011} & 0.041                                                             & 0.018                                                              & 0.016                                                           & 0.052                                                            & 0.016                                                           \\ 
			& 40\%                                                                & \textbf{0.011} & 0.051                                                             & 0.017                                                              & 0.029                                                           & 0.113                                                            & 0.029                                                           \\ \hline
			\multirow{3}{*}{Digits}    & 20\%       & \textbf{0.058} & 0.071      & 0.060    & 0.070    & 0.071   & 0.070   \\
			& 30\%      & \textbf{0.058} & 0.070   & 0.064      & 0.071    & 0.082  & 0.072 \\ 
			& 40\%    & \textbf{0.074} & 0.084     & 0.081    & 0.084         & 0.106       & 0.082   \\ \hline
			\multirow{3}{*}{Inosphere} & 20\% & \textbf{0.155} & 0.159 & 0.161 & 0.159 &  0.169 & 0.159 \\
			& 30\% & 0.139 & 0.151 &\textbf{ 0.134} & 0.146 & 0.159 & 0.148 \\
			& 40\% & \textbf{0.151} & 0.153 & 0.164 & 0.170 & 0.191 & 0.165 \\ \hline
	\end{tabular}}
		\caption{The cross-validation errors on datasets with different missing rates in LDA application}
\end{table*}

\begin{figure}
	\centering
	\includegraphics[width=0.6\textwidth]{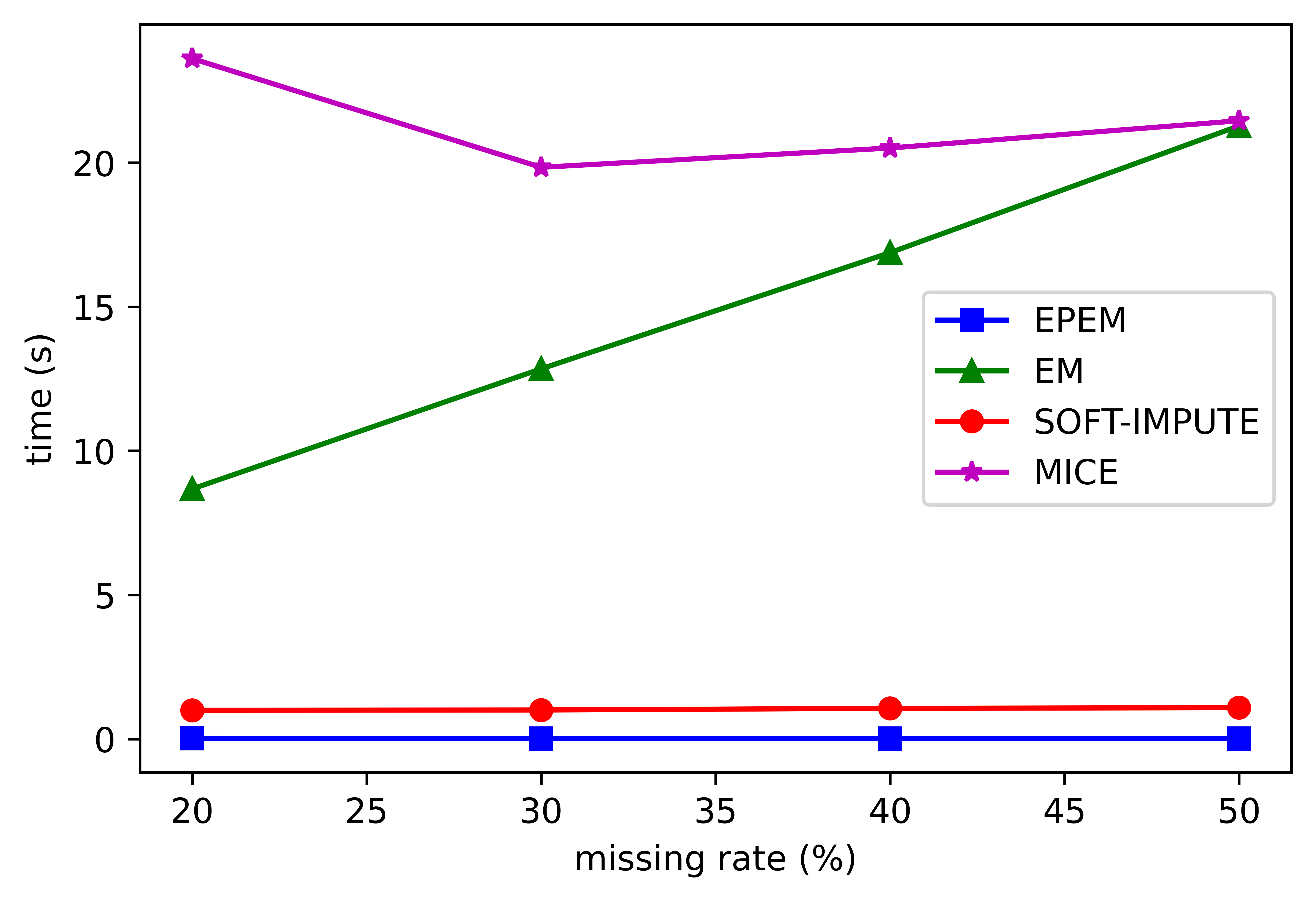}
	\caption{The running time w.r.t different missing rates.}
	\label{fig1}
\end{figure}

\subsection{Results and Discussion}

From Table \ref{paraerr}, one can see that EPEM consistently outperforms other approaches by a significant margin. For example, with $40\%$ of missing data in the Parkinson data set, EPEM gives the error rate of $0.043$, which is less than $1/2$ the error rate of other approaches (including powerful methods such as e.g., SOFT-IMPUTE, MICE, KNN, EM, and NNMT). Another example can be seen in the Digits dataset, with $20\%$ missing rate, the error rate of EPEM ($0.003$) is approximate $\frac{1}{10}$ of the next best results ($0.026$). Note that the Royston test for testing multivariate normality gives $p-$value $<0.01$ for all of the datasets in this experiment, which indicates that all those datasets do not follow a multivariate normal distribution. Therefore, even though multivariate normality is assumed for the proof of all the theorems in this paper, the results are robust to normality.

When our estimation scheme is utilized for LDA, one can see from Table \ref{cverr} that the proposed technique also outperforms other methods over most of the data sets in the experiment again. For the Iris dataset with a $20\%$ missing rate, our approach surpasses MICE imputation by $3.9\%$ error rate reduction, SOFT-IMPUTE by $9.3\%$, and KNN by $10.1\%$. Another example is for the Parkinson dataset with a $30\%$ missing rate, and our method exceeds KNN imputation and Nuclear norm minimization by reducing $3.5\%$, $4.9\%$ the error rate, respectively. 

Related to the running time, as depicted in Figure \ref{fig1}, our algorithm surpasses all the algorithms in the efficiency. It is essential to note that the running time for both KNN and Nuclear Norm minimization is discarded from the figure for a more unobstructed view of the remaining algorithm (the minimum run time for Nuclear Norm minimization $424.43$ seconds, and for KNN is $109.47$ seconds). This advantage benefits from the exact formula in our algorithm instead of running several iteration steps as other approaches.

\section{Conclusion}
In this paper, we have derived the exact formulas and proposed an efficient algorithm for computing the parameters of a multiple class monotone missing data set, where the covariance matrices of all the classes are assumed to be equal. Even though normality is assumed, we have illustrated that the algorithm 
works well in practice and can achieve good performance through different experiments compared to several other imputation based approaches. While we only illustrate with two typical applications, our method is general enough to extend for other scenarios such as principle component analysis, Fisher's discriminant analysis and hypothesis testing for multiple class monotone missing data. 
All codes and data in our experiments are available at the following repository: \url{https://github.com/thunguyen177/EPEM}.
\bibliographystyle{unsrt}
\bibliography{ref}

\newpage
\appendix

\section{An illustration for data partition on Section 3.1}
\label{sec:appendix_A}
If one considers the following data set
\begin{equation*}
\begin{pmatrix} 
m_{11} & m_{12} & m_{13} & m_{14} & m_{15}\\
m_{21} & m_{22} & m_{23} & m_{24} & m_{25}\\
m_{31} & m_{32} & * & * & *\\
m_{41} & m_{42} & * & * & *
\end{pmatrix},
\end{equation*}
in which, 
\begin{equation*}
\boldsymbol{x}=\begin{pmatrix} 
m_{11} & m_{13}\\
m_{21} & m_{23}\\
m_{31} & *\\
m_{41} & *
\end{pmatrix},\;\;\;
\boldsymbol{y}=\begin{pmatrix} 
m_{12} & m_{14} & m_{15}\\
m_{22} & m_{24} & m_{25}\\
m_{32} & * &*\\
m_{42} & * & *
\end{pmatrix}.
\end{equation*}
Then, 
\begin{equation*}
\boldsymbol{x}_1 = \begin{pmatrix}
m_{11}&m_{13}\\
m_{21}&m_{23}
\end{pmatrix},
\boldsymbol{x}_2 = \begin{pmatrix}
m_{11}\\m_{21}\\m_{31}\\m_{41}
\end{pmatrix},\;\;\;
\boldsymbol{y}_1 = \begin{pmatrix}
m_{12} & m_{14} & m_{15}\\
m_{22} & m_{24} & m_{25}
\end{pmatrix},
\boldsymbol{y}_2 = \begin{pmatrix}
m_{12} \\ m_{22} \\ m_{32}\\m_{42}
\end{pmatrix}.
\end{equation*}

\section{Proof of Theorem 3.1}
\label{sec:appendix_B}
This is a special case of a general problem for multiple groups, of which, the proof will be given in the next sections.

\section{Proof of Theorem 3.2}
\label{sec:appendix_C}

The proof makes use of the following identities from \cite{petersen2012matrix}: 

\textit{For a vector $\boldsymbol{x}$ and matrices $\mathbf{A},\mathbf{B}, \mathbf{C}$ of appropriate sizes
	\begin{align}
	\frac{\partial \boldsymbol{x}'\mathbf{B}\boldsymbol{x}   }{\partial \boldsymbol{x} }&=(\mathbf{B}+\mathbf{B}')\boldsymbol{x} \label{a}\\
	\frac{\partial tr(\mathbf{A}\mathbf{X} \mathbf{B} )}{\partial \mathbf{X} } &= \mathbf{A}'\mathbf{B} '\label{b}\\ 
	\frac{\partial tr(A\mathbf{X}\mathbf{B}\mathbf{X}'\mathbf{C} )}{\partial \mathbf{X} } &= \mathbf{A}'\mathbf{C}'\mathbf{X}\mathbf{B}'+\mathbf{C} \mathbf{A}\mathbf{X}\mathbf{B} \label{c},
	\end{align}  
	where $A'$ denotes the transpose of a matrix $A$.}

Now, we will prove this theorem. First, notice that maximizing the likelihood w.r.t $\boldsymbol {\mu }$ is equivalent to maximizing the likelihood w.r.t  $\boldsymbol {\mu } ^{(g)}, g=1,..., G$. We partition $\boldsymbol{\mu}^{(g)}$ as
\begin{equation} 
\boldsymbol{\mu}^{(g)}=  \begin{pmatrix}
[\boldsymbol{\mu}^{(g)}]_1\\
[\boldsymbol{\mu}^{(g)}]_2\\
\vdots\\
[\boldsymbol{\mu}^{(g)}]_k
\end{pmatrix} \text{ of sizes } \begin{pmatrix}
p_1 \times 1\\
p_2 \times 1\\
\vdots\\
p_k \times 1
\end{pmatrix}, g=1,..., G.
\end{equation}

Then, let

\begin{equation}\label{part-mu}
\boldsymbol{\mu}^{(g)}_i=  \begin{pmatrix}
[\boldsymbol{\mu}^{(g)}]_1\\
[\boldsymbol{\mu}^{(g)}]_2\\
\vdots\\
[\boldsymbol{\mu}^{(g)}]_i
\end{pmatrix}, g=1,..., G,\; i = 1,...k.
\end{equation}

To represent $\widehat{[\boldsymbol {\mu }^{(g)}]}_i$ in terms of $\widehat{\boldsymbol {\mu }}_{i-1}^{(g)}$, one can also partition

\begin{equation}\label{lab5}
\bar{\boldsymbol{x}}^{(g)}_{i} = \begin{pmatrix}
{[\bar{\boldsymbol{x}}^{(g)}_{i}]_1}\\
{[\bar{\boldsymbol{x}}^{(g)}_{i}]_2}
\end{pmatrix} \text{ of sizes } 
\begin{pmatrix}
{(\sum_{j=1}^{i-1}p_j)\times 1}\\
{p_i\times 1}
\end{pmatrix}.
\end{equation}


We assume that $\boldsymbol{z}_{ij}^{(g)}$ be the $j^{th}$ column of $\boldsymbol{x}_i^{(g)}$.
For $h\ge 2$, we denote $[\boldsymbol{z}_{ij}^{(g)}]_{h-1}$ as a vector containing the first $\sum_{s=1}^{h-1}p_s$ elements of $\boldsymbol{z}_{ij}^{(g)}$ and $[\boldsymbol{z}_{ij}^{(g)}]_{h/(h-1)}$ includes the next $p_h$ elements. 

Let $n_{k+1}^{(g)} = 0$ for all $1\le g\le G$, and use $f$ to denote the density of the variable inside $"f(.)"$ in general (for notation simplicity). In addition, denote by $N_d(\boldsymbol {\gamma}, \boldsymbol{x}) $ the $d-$dimensional multivariate normal distribution with mean $\boldsymbol {\gamma}$ and covariance matrix $\boldsymbol {\Sigma }$. We apply the following Lemma \ref{lm:1} to prove Theorem 2.

\begin{lemma}
	The log likelihood function  can be formulated as
	\begin{equation}
	\eta  = \sum_{i=1}^k \eta_i,
	\end{equation}
	where 
	\begin{equation}
	\eta_1 = \sum_{g=1}^{G}  \sum_{l=1}^k \sum_{j=n_{l+1}^{(g)}+1}^{n_l^{(g)}} f(\boldsymbol{z}_{1j}^{(g)}),
	\end{equation}
	and 
	\begin{equation}
	\eta_i = \sum_{g=1}^{G} \sum_{l=i}^k \sum_{j=n_{l+1}^{(g)}+1}^{n_l^{(g)}} 
	f([\boldsymbol{z}_{lj}^{(g)}]_{i/(i-1)}|[\boldsymbol{z}_{lj}^{(g)}]_{i-1}), \;\;\; 2\le i \le k.
	\end{equation}
	\label{lm:1}
\end{lemma}

\textbf{(a) To derive MLEs of $\boldsymbol {\mu } ^{(g)}_1, \boldsymbol {\Sigma }_1$}:

Deriving the MLEs of $\boldsymbol {\mu } ^{(g)}_1, \boldsymbol {\Sigma }_1$ is straightforward by noting that $\boldsymbol{z}_{1g}^{(g)}\sim N_{p_1}(\boldsymbol {\mu }_1^{(g)}, \boldsymbol {\Sigma }_1)$ as well as
\begin{equation}
\frac{\partial \eta}{\partial \boldsymbol {\mu } ^{(g)}_1} = 	\frac{\partial \eta_1}{\partial \boldsymbol {\mu } ^{(g)}_1} \;\;\; \text{and}\;\;\;
\frac{\partial \eta}{\partial \boldsymbol {\Sigma }_1} = 	\frac{\partial \eta_1}{\partial \boldsymbol {\Sigma }_1},
\end{equation}
and solving for $\boldsymbol {\mu } ^{(g)}_1$ and $\boldsymbol {\Sigma }_1$ from 
\begin{equation}
\frac{\partial \eta_1}{\partial \boldsymbol {\mu } ^{(g)}_1} =0 \;\;\; \text{and} \;\;\;	\frac{\partial \eta_1}{\partial \boldsymbol {\Sigma }_1} = 0.
\end{equation}

\textbf{(b) Explicit expression for $\eta_i (2\le i \le k)$:}
To find an explicit expression for $\eta_i$, we use the following lemma, whose proof could be found in \cite{johnson2002applied}:

\textbf{Lemma 2} \textit{(conditional distribution for multivariate normal distribution) Suppose $\mathbf{y} \sim N_p(\boldsymbol {\gamma }, \boldsymbol {\Sigma })$. Let }
\begin{equation*}
{\mathbf  {y}}={\begin{bmatrix}{\mathbf  {y}}_{1}\\{\mathbf  {y}}_{2}\end{bmatrix}}{\text{ with sizes }}{\begin{bmatrix}q\times 1\\(p-q)\times 1\end{bmatrix}}
\end{equation*}
\textit{and accordingly we partition
}\begin{equation*}
{\boldsymbol {\gamma }}={\begin{bmatrix}{\boldsymbol {\gamma }}_{1}\\{\boldsymbol {\gamma }}_{2}\end{bmatrix}}{\text{ with sizes }}{\begin{bmatrix}q\times 1\\(p-q)\times 1\end{bmatrix}}
\end{equation*}
\textit{and} 
\begin{equation*}
\boldsymbol\Sigma
=
\begin{bmatrix}
\boldsymbol\Sigma_{11} & \boldsymbol\Sigma_{12} \\
\boldsymbol\Sigma_{21} & \boldsymbol\Sigma_{22}
\end{bmatrix}
\text{ with sizes }\begin{bmatrix} q \times q & q \times (p-q) \\ (p-q) \times q & (p-q) \times (p-q) \end{bmatrix}.
\end{equation*}
\textit{Then the distribution of $\mathbf{y}_1$ conditional on $\mathbf{y}_2=a$ follows multivariate normal distribution with mean
}\begin{equation*}
\bar{\boldsymbol\gamma}
=
\boldsymbol\gamma_1 + \boldsymbol\Sigma_{12} \boldsymbol\Sigma_{22}^{-1}
\left(
\mathbf{a} - \boldsymbol\gamma_2
\right)
\end{equation*}
\textit{and covariance matrix
}\begin{equation*}
{\overline {\boldsymbol {\Sigma }}}={\boldsymbol {\Sigma }}_{11}-{\boldsymbol {\Sigma }}_{12}{\boldsymbol {\Sigma }}_{22}^{-1}{\boldsymbol {\Sigma }}_{21}.
\end{equation*}

To continue with our proof, recall that for $i\ge 2$, $[\boldsymbol{z}_{lj}^{(g)}]_{i-1}$ contains the first $\sum_{s=1}^{i-1}p_s$ elements of $\boldsymbol{z}_{lj}^{(g)}$ and $[\boldsymbol{z}_{lj}^{(g)}]_{i/(i-1)}$ contains the next $p_i$ elements. Hence
\begin{equation*}
[\boldsymbol{z}_{lj}^{(g)} ]_i= \begin{pmatrix}
[\boldsymbol{z}_{lj}^{(g)} ]_{i-1}\\
[\boldsymbol{z}_{lj}^{(g)} ]_{i/(i-1)} 
\end{pmatrix}.
\end{equation*}
In addition,
\begin{equation*}
[\boldsymbol{z}^{(g)}_{lj}]_i\sim N_{\sum_{j=1}^i p_j}(\boldsymbol{\mu}_i^{(g)}, \bSigma_i).
\end{equation*}
Therefore, 
we can prove similar to the lemma above that
\begin{equation}\label{lab12}
[\boldsymbol{z}_{lj}^{(g)}]_{i/(i-1)}|[\boldsymbol{z}_{lj}^{(g)}]_{i-1} \sim N_{p_i}\left(
\mathbf{m} ,\mathbf{H} \right),
\end{equation}
where
\begin{align}	
\mathbf{m}  &=
[\boldsymbol{\mu}^{(g)}]_i+B_i([\boldsymbol{z}_{lj}^{(g)}]_{i-1}-\boldsymbol{\mu}^{(g)}_{i-1}),\\
\mathbf{H}  &= [\bSigma_i]_{22}-B_i[\bSigma_i]_{12},\label{h}
\end{align}
and 
\begin{equation}\label{bi}
B_i=[\bSigma_{i}]_{21}([\bSigma_i]_{11})^{-1}.
\end{equation}
(We ignored the indices of $\mathbf{m} $ and $\mathbf{H} $ for notational simplicity.) 

Therefore, maximizing the likelihood w.r.t. $[\boldsymbol{\mu}^{(g)}]_i$,$ [\bSigma_i]_{12},[\bSigma_i]_{11}$ is equivalent to maximizing it w.r.t $[\boldsymbol{\mu}^{(g)}]_i$, $B_i, \mathbf{H} $. From Eq. \eqref{lab12} we have
\begin{eqnarray}\label{etai}
\eta_i = C -\frac{1}{2}\sum_{g=1}^G \sum_{l=i}^{k}(n_l-n_{l+1})\ln \mid\mathbf{H} \mid
- \frac{1}{2}\sum_{g=1}^G \sum_{l=i}^{k}\sum_{j=n_{l+1}^{(g)}+1}^{n_l^{(g)}} 
u_{iglj}'\mathbf{H} ^{-1}u_{iglj},
\end{eqnarray}
where  $C$ is a constant and $ 2\le i \le k$.
\begin{equation}\label{uiglj}
u_{iglj} = [\boldsymbol{z}_{lj}^{(g)}]_{i/(i-1)} - [\boldsymbol{\mu}^{(g)}]_i- B_i\left([\boldsymbol{z}_{lj}^{(g)}]_{i-1}-\boldsymbol{\mu}^{(g)}_{i-1}\right).
\end{equation}
\textbf{(c) To derive MLEs of $[\boldsymbol {\mu} ^{(g)}]_{i}(2\le i \le k)$:}

Using Eq. (\ref{a}), we have
\begin{equation*}
\frac{\partial \eta_i}{\partial u_{iglj}} = 
-\frac{1}{2}2\mathbf{H} ^{-1}u_{iglj} \;\;\text{and}\;\;
\partial u_{iglj} = - \partial [\boldsymbol{\mu}^{(g)}]_i.
\end{equation*}
Therefore,
\begin{equation*}
\frac{\partial \eta}{\partial u_{iglj}} =\frac{\partial \eta_i}{\partial [\boldsymbol{\mu}^{(g)}]_i} = 
\mathbf{H} ^{-1}\sum_{l=i}^{k}\sum_{j=n_{l+1}^{(g)}+1}^{n_l^{(g)}} u_{iglj},
\end{equation*}
which implies
\begin{align}
\small
\frac{\partial \eta}{\partial [\boldsymbol{\mu}^{(g)}]_i} = 0 & \Longleftrightarrow  \sum_{l=i}^{k}\sum_{j=n_{l+1}^{(g)}+1}^{n_l^{(g)}} u_{iglj} = 0\\
&\overset{Eq. (\ref{uiglj})}{\Longleftrightarrow}  \sum_{l=i}^{k}\sum_{j=n_{l+1}^{(g)}+1}^{n_l^{(g)}} 
\left[[\boldsymbol{z}_{lj}^{(g)}]_{i/(i-1)} - [\boldsymbol{\mu}^{(g)}]_i- B_i\left([\boldsymbol{z}_{lj}^{(g)}]_{i-1}-\boldsymbol{\mu}^{(g)}_{i-1}\right)\right]= 0\label{new}
\end{align}
In addition, note that $\bar{\boldsymbol{x}}_i^{(g)}$ is the mean of $\boldsymbol{x}^{(g)}_i$. Therefore,
\begin{equation}
\bar{\boldsymbol{x}}_i^{(g)} = \frac{\sum_{l=i}^{k}\sum_{j=n_{l+1}^{(g)}+1}^{n_l^{(g)}} \boldsymbol{z}_{lj}^{(g)} }{\sum_{l=i}^{k}(n_l^{(g)}-n_{l+1}^{(g)})}.
\end{equation}
Therefore, by simplifying Eq. \eqref{new}, using Eq. \eqref{lab5} with the above fact yields an estimation for $[\boldsymbol{\mu}^{(g)}]_i$:
\begin{equation}\label{muh}
[\widehat{\boldsymbol{\mu}}^{(g)}]_i = [\bar{\boldsymbol{x}}^{(g)}_{i}]_2-B_i([\bar{\boldsymbol{x}}^{(g)}_{i}]_1-\widehat{\boldsymbol{\mu}}^{(g)}_{i-1}).
\end{equation}

\textbf{(d) To derive the MLEs of $\mathbf{B}_i (i = 2,\dots,k)$}:

Plugging Eq. (\ref{muh}) into $\eta_i$, we have 
\begin{equation}\label{et2}
\eta_i
= C -\frac{1}{2}\sum_{g=1}^G \sum_{l=i}^{k}(n_l^{(g)}-n_{l+1}^{(g)})\ln \mid\mathbf{H} \mid-\frac{1}{2} tr \left( \mathbf{H} ^{-1}D\right),
\end{equation}
where $tr(A)$ is the trace of matrix $A$ and 
\begin{align*}
D =\sum_{g=1}^G \sum_{l=i}^{k}\sum_{j=n_{l+1}^{(g)}+1}^{n_l}
&\left[\left(([\boldsymbol{z}_{lj}^{(g)}]_{i/(i-1)} -[\bar{\boldsymbol{x}}^{(g)}_{i}]_2)\right.
-B_i\left([\boldsymbol{z}_{lj}^{(g)}]_{i-1}-[\bar{\boldsymbol{x}}^{(g)}_{i}]_1\right)\right)\\
&\times\left(([\boldsymbol{z}_{lj}^{(g)}]_{i/(i-1)} -[\bar{\boldsymbol{x}}^{(g)}_{i}]_2)  -\left.B_i\left([\boldsymbol{z}_{lj}^{(g)}]_{i-1}-[\bar{\boldsymbol{x}}^{(g)}_{i}]_1\right)\right)'\right].
\end{align*}
Moreover, recall that 
\begin{equation}
\mathbf{W} _i = \sum_{g=1}^G (n_i^{(g)}-1)S_i^{(g)},\; i = 1,...,k.
\end{equation}
Therefore, by factoring, we can get
\begin{equation*}
D =  [\bW_i]_{22} -B_i[\bW_i]_{12}- (B_i [\bW_i]_{12})'+B_i[\bW_i]_{11}B_i'.
\end{equation*}
Hence, taking derivative of $\eta_i$ in Eq. \eqref{et2} w.r.t $B_i$, we have
\begin{align*}
\frac{\partial \eta_i}{\partial B_i} \propto 
\frac{\partial }{\partial B_i}&\left[
tr ( \mathbf{H} ^{-1}[\bW_i]_{22})-
tr ( \mathbf{H} ^{-1}B_i[\bW_i]_{12})\right.\\
&-\left.tr \left( \mathbf{H} ^{-1}(B_i[\bW_i]_{12})'\right)+
tr ( \mathbf{H} ^{-1}B_i[\bW_i]_{11}B_i') 
\right].
\end{align*}
Moreover, since the trace of a matrix is equal to the trace of its transpose, it implies that
\begin{align*}
tr( \mathbf{H} ^{-1}(B_i[\bW_i]_{12})') &= tr(B_i[\bW_i]_{12}\mathbf{H} ^{-1}) \\
&= 	tr ( \mathbf{H} ^{-1}B_i[\bW_i]_{12}).
\end{align*}
Therefore, using identities \eqref{b} and \eqref{c}, we have
\begin{equation*}
\frac{\partial \eta}{\partial B_i}=\frac{\partial \eta_i}{\partial B_i} \propto  
-2\mathbf{H} ^{-1}[\bW_i]_{12}' + \mathbf{H} ^{-1}B_i[\mathbf{W}  _i]_{11}+ \mathbf{H} ^{-1}B_i[\mathbf{W}  _i]_{11}.    
\end{equation*}
Hence, 
\begin{equation*}
\frac{\partial \eta}{\partial B_i} = 0 \Longleftrightarrow
B_i[\mathbf{W}  _i]_{11} = [\bW_i]_{12}'=[\bW_i]_{21},
\end{equation*}
which means the maximum likelihood estimate of $B_i$ is
\begin{equation}\label{pi}
\bP_i =  [\bW_i]_{21}  [\bW_i]_{11} ^{-1}.
\end{equation}

\textbf{(e) To estimate $\mathbf{H}$:}

Computing the MLE for $\mathbf{H} $ is similar to finding the MLE for the covariance matrix of a multivariate normal distribution: Taking derivative of $\eta_i$ in Eq. \eqref{et2} w.r.t $\left(\mathbf{H} \right)^{-1}$ and using the trace trick $\boldsymbol{b}'A\boldsymbol{b} = tr(A\boldsymbol{b}\boldsymbol{b}')$ for a matrix A and a vector $\boldsymbol{b}$, we can obtain the  maximum likelihood estimate of $\mathbf{H} $ as follows:
\begin{equation}\label{qi}
\bQ_i =\frac{1}{\sum_{g=1}^G n_i^{(g)}}\left([\bW_i]_{22}-\bP_i [\bW_i]_{12} \right).
\end{equation}
\textbf{(f) To compute $\widehat{[\bSigma_i]}_{22}$ and $\widehat{[\bSigma_i]}_{12}$}:

From Eq. \eqref{bi} and Eq. \eqref{pi}, one can see that the estimate for $[\bSigma_i]_{12}$ can be given by:
\begin{equation}
\widehat{[\bSigma_i]}_{12} = P_i\widehat{[\bSigma_i]}_{11}
\end{equation}

Next, from Eq. \eqref{h} and Eq. \eqref{qi}, we can calculate the estimate for ${[\bSigma_i]}_{22}$ as:
\begin{equation}
\widehat{[\bSigma_i]}_{22} = Q_i + P_i \widehat{[\bSigma_i]}_{12}
\end{equation}

\textbf{(g) To prove that} $[\hat{\boldsymbol {\mu }}]_i=[\bar{\boldsymbol{x}}_i]_2 - \mathbf{P}  _i([\bar{\boldsymbol{x}}_i^{(g)}]_1-\hat{\boldsymbol {\mu }}_{i-1})$:	

Recall that $\widehat{\boldsymbol {\mu }}_1 = \bar{\boldsymbol{x}}_1$ and plugging Eq. \eqref{bi} into Eq. \eqref{muh} gives
\begin{equation}
[\widehat{\boldsymbol {\mu }}^{(g)}]_i = [\bar{\boldsymbol{x}}_i^{(g)}]_2 - \mathbf{P}_i([\bar{\boldsymbol{x}}_i^{(g)}]_1-\widehat{\boldsymbol {\mu }}_{i-1}^{(g)}).
\end{equation}
In addition, recall that
\begin{align*}
\widehat{\boldsymbol {\mu } }_i =\begin{pmatrix}
\widehat{\boldsymbol {\mu }}_{i-1}\\
[\widehat{\boldsymbol {\mu }}]_i
\end{pmatrix} &= \begin{pmatrix}
\widehat{\boldsymbol {\mu }}^{(1)}_{i-1}, \widehat{\boldsymbol {\mu }}^{(2)}_{i-1},...,\widehat{\boldsymbol {\mu }}^{(G)}_{i-1}\\
[\widehat{\boldsymbol {\mu }}^{(1)}]_i, [\widehat{\boldsymbol {\mu }}^{(2)}]_i,...,[\widehat{\boldsymbol {\mu }}^{(G)}]_i
\end{pmatrix},\\
\bar{\boldsymbol{x}}_i =\begin{pmatrix}
[\bar{\boldsymbol{x}}_i]_1\\
[\bar{\boldsymbol{x}}_i]_2
\end{pmatrix} &= \begin{pmatrix}
[\bar{\boldsymbol{x}}_i^{(1)}]_1, [\bar{\boldsymbol{x}}_i^{(2)}]_1,...,[\bar{\boldsymbol{x}}_i^{(G)}]_1\\
[\bar{\boldsymbol{x}}_i^{(1)}]_2, [\bar{\boldsymbol{x}}_i^{(2)}]_2,...,[\bar{\boldsymbol{x}}_i^{(G)}]_2
\end{pmatrix}.    
\end{align*}
Moreover, note that for matrices $E,F$ of order ${m\times n}, {n\times s}$, respectively, and if $f_1,.., f_s$ are the $1^{st},2^{nd},...,s^{th}$ columns of  F then $EF = (Ef_1,Ef_2,...,Ef_s)$. Therefore, 
\begin{equation*}
[\hat{\boldsymbol {\mu }}]_i=[\bar{\boldsymbol{x}}_i]_2 - \mathbf{P}  _i([\bar{\boldsymbol{x}}_i^{(g)}]_1-\hat{\boldsymbol {\mu }}_{i-1}).
\end{equation*}
\subsection{Proof of  Lemma \ref{lm:1}}\label{likeli}
The proof makes use of the following identity: 
\begin{equation}\label{ali}
\prod_{l=2}^{k}\prod_{i=2}^{l} a_{li}=\prod_{i=2}^{k}\prod_{l=i}^{k} a_{li}
\end{equation}
for any sequence $\{a_{li}\}_{l=2, i=2}^{k,l}$.

\textbf{Proof of the identity}
The proof of the identity is straight forward by noting that
\begin{equation}
a_{22}(a_{32} a_{33})(a_{42}a_{43}a_{44})....(a_{k1}a_{k2}...a_{kk})
= (a_{22}a_{32}...a_{k2})(a_{33}a_{43}...a_{k3})....a_{kk}
\end{equation}
\textbf{Proof of  Lemma \ref{lm:1}}
Recall that $\boldsymbol{z}_{lj}^{(g)}$ is the $j^{th}$ column of $\boldsymbol{x}_l^{(g)}$. Therefore, 
the likelihood is 
\begin{equation}
L = \prod_{g=1}^{G}\prod_{l=1}^k \prod_{j=n_{l+1}^{(g)}+1}^{n_l^{(g)}}f(\boldsymbol{z}_{lj}^{(g)})=\prod_{g=1}^{G} L^{(g)},
\end{equation}
where 
\begin{equation}
L^{(g)}= \prod_{l=1}^k \prod_{j=n_{l+1}^{(g)}+1}^{n_l^{(g)}}f(\boldsymbol{z}_{lj}^{(g)}).
\end{equation}
Recall that for $2\le h\le k$, $[\boldsymbol{z}_{lj}^{(g)}]_{h-1}$ contains the first $\sum_{s=1}^{h-1}p_s$ elements of $\boldsymbol{z}_{lj}^{(g)}$ and $[\boldsymbol{z}_{lj}^{(g)}]_{h/(h-1)}$ contains the next $p_h$ elements. Hence
\begin{equation}
[\boldsymbol{z}_{lj}^{(g)} ]_h= \begin{pmatrix}
[\boldsymbol{z}_{lj}^{(g)} ]_{h-1}\\
[\boldsymbol{z}_{lj}^{(g)} ]_{h/(h-1)} 
\end{pmatrix}
\end{equation}
and
\begin{align}
[\boldsymbol{z}_{lj}^{(g)}]_1 &= \boldsymbol{z}_{1j}^{(g)}\;\; \forall i=1,..,k\label{lab1}\\
[\boldsymbol{z}_{lj}^{(g)}]_l &= \boldsymbol{z}_{lj}^{(g)}\;\; \forall l=2,..,k. \label{lab2}
\end{align}
Therefore, for $2\le l\le k$:
\begin{equation*}	f(\boldsymbol{z}_{lj}^{(g)})=f(\boldsymbol{z}_{1j}^{(g)})
\frac{f([\boldsymbol{z}_{lj}^{(g)}]_2)}{f(\boldsymbol{z}_{1j}^{(g)} )}
\frac{f([\boldsymbol{z}_{lj}^{(g)}]_3)}{f([\boldsymbol{z}_{lj}^{(g)}]_2 )}...
\frac{f([\boldsymbol{z}_{lj}^{(g)}]_l)}{f([\boldsymbol{z}_{lj}^{(g)}]_{l-1} )}.
\end{equation*}
Next, note that
\begin{equation*}
\frac{f([\boldsymbol{z}_{lj}^{(g)}]_s)}{f([\boldsymbol{z}_{lj}^{(g)}]_{s-1} )}	= f([\boldsymbol{z}_{lj}^{(g)}]_{s/(s-1)}|[\boldsymbol{z}_{lj}^{(g)}]_{s-1}).
\end{equation*}

Hence, combining with Eq. (\ref{lab1}) and Eq. (\ref{lab2}) implies

\begin{align*}
f(\boldsymbol{z}_{lj}^{(g)}) &= f(\boldsymbol{z}_{1j}^{(g)}) f([\boldsymbol{z}_{lj}^{(g)}]_{2/1}|\boldsymbol{z}_{1j}^{(g)})
f([\boldsymbol{z}_{lj}^{(g)}]_{3/2}|[\boldsymbol{z}_{lj}^{(g)} ]_2)
...f([\boldsymbol{z}_{lj}^{(g)}]_{l/(l-1)}|[\boldsymbol{z}_{lj}^{(g)} ]_{l-1})\\
&=f(\boldsymbol{z}_{1j}^{(g)})\prod_{i=2}^l f([\boldsymbol{z}_{lj}^{(g)}]_{i/(i-1)}|[\boldsymbol{z}_{l,j}^{(g)}]_{i-1}).
\end{align*}
This deduces that 
\begin{align*}
L^{(g)}= &\left[\prod_{l=2}^k \prod_{j=n_{l+1}^{(g)}+1}^{n_l^{(g)}}\left( f(\boldsymbol{z}_{1j}^{(g)}) 
\prod_{i=2}^l f([\boldsymbol{z}_{lj}^{(g)}]_{i/(i-1)}|[\boldsymbol{z}_{lj}^{(g)})]_{i-1} \right) \right]  
\times\prod_{j=n_2^{(g)}+1}^{n_1^{(g)}} f(\boldsymbol{z}^{(g)}_{1j})\\
&= \left( \prod_{l=1}^k \prod_{j=n_{l+1}^{(g)}+1}^{n_l^{(g)}}f(\boldsymbol{z}_{1j}^{(g)}) \right) \times
\left(\prod_{l=2}^k \prod_{j=n_{l+1}^{(g)}+1}^{n_l^{(g)}} \prod_{i=2}^l f([\boldsymbol{z}_{lj}^{(g)}]_{i/(i-1)}|[\boldsymbol{z}_{lj}^{(g)})]_{i-1} \right).
\end{align*}
Together with Eq. \eqref{ali}, this yields,
\begin{align*}
L &= \left(\prod_{g=1}^{G}  \prod_{l=1}^k \prod_{j=n_{l+1}^{(g)}+1}^{n_l^{(g)}}f(\boldsymbol{z}_{1j}^{(g)}) \right) \times
\left(\prod_{i=2}^k\prod_{g=1}^{G} \prod_{l=i}^k \prod_{j=n_{l+1}^{(g)}+1}^{n_l^{(g)}}  f([\boldsymbol{z}_{lj}^{(g)}]_{i/(i-1)}|[\boldsymbol{z}_{lj}^{(g)}]_{i-1}) \right)\\
& = \prod_{i=1}^k L_i,
\end{align*}	
where 
\begin{equation}
L_1 = \prod_{g=1}^{G}  \prod_{l=1}^k \prod_{j=n_{l+1}^{(g)}+1}^{n_l^{(g)}}f(\boldsymbol{z}_{1j}^{(g)}),
\end{equation}
and 
\begin{equation}
L_i = \prod_{g=1}^{G} \prod_{l=i}^k \prod_{j=n_{l+1}^{(g)}+1}^{n_l^{(g)}} f([\boldsymbol{z}_{lj}^{(g)}]_{i/(i-1)}|[\boldsymbol{z}_{lj}^{(g)}]_{i-1}), \;\;\; 2\le i \le k.
\end{equation}
Therefore, the log likelihood can be calculated as:
\begin{equation*}
\eta  = \sum_{i=1}^k\log L_i =  \sum_{i=1}^k \eta_i.
\end{equation*}

\section{Proof of Theorem 3.3}
\label{sec:appendix_D}
Maximum likelihood estimates are asymptotic (see \cite{casella2002statistical}). Therefore, the proof follows directly from the result of Theorem 3.2.

\section{Proof of Theorem 4.1}
\label{sec:appendix_E}

The proof is straightforward by using the well-known result that if $\widehat {\theta \,}$ is the MLE for $\theta$, and if $g(\theta)$ is any transformation of $\theta$ , then the MLE for $g(\theta )$ is $g(\,{\widehat {\theta \,}}\,)$ (see \cite{casella2002statistical}). 

Let $\by=A\boldsymbol{x}$ and suppose that $\boldsymbol{x}$ belongs to the $i$th class, then the new data follow a normal distribution with mean $A\mu_i$ and covariance matrix $A\Sigma A'$. The MLEs for the mean and covariance matrix of the transformed data are $A\hat{\boldsymbol{\mu}}_i, A\widehat{\bSigma}A'$, respectively. The linear discriminant analysis classification rule after the transformation can be given as:
\begin{equation} 
\hat{d}_i(\by) = (A\hat{\boldsymbol{\mu}}_i)'\widehat{\bSigma}^{-1}A^{-1}\hat{\boldsymbol{x}}-\frac{1}{2}(A\hat{\boldsymbol{\mu}}_i)'(A\widehat{\bSigma}A')^{-1}A\hat{\boldsymbol{\mu}}_i+\ln \hat{r}_i, \;\;
\end{equation}
for $ i=1,2,...,G.$

As $A$ is invertible, the above equation simplifies to
\begin{equation}
\hat{d}_i(\boldsymbol{x}) = \hat{\boldsymbol{\mu}}_i'\widehat{\bSigma}^{-1}\hat{\boldsymbol{x}}-\frac{1}{2}\hat{\boldsymbol{\mu}}_i'\widehat{\bSigma}^{-1}\hat{\boldsymbol{\mu}}_i+\ln \hat{r}_i, 
\end{equation}
and the proof follows.

\section{Proof of Theorem 4.2}
\label{sec:appendix_F}
\begin{lemma}{(Slutsky's theorem)}
	Let $X_n,X$ be random vectors and $Y_n$ be random matrices. If $X_n\xrightarrow{d}X$ and $Y_n\xrightarrow{d}c$ where $c$ is a constant matrix, then
	\begin{itemize}
		\item[i)] $X_nY_n\xrightarrow{d}Xc$;
		\item[ii)] $X_nY^{-1}_n\xrightarrow{d}Xc^{-1}$, provided $Y_n$ and $c$ are invertible matrices.
	\end{itemize}
\end{lemma}
The proof of this lemma can be found in \cite{van2000asymptotic}.

Now, let us fix the value of $\boldsymbol{x}$. We will prove for case $i=1$, and then other cases can be done similarly.

First, $\hat{\boldsymbol{\mu}}_1$ and $\widehat{\bSigma}$ are the maximum likelihood estimators of $\boldsymbol{\mu}_1$ and $\bSigma$, respectively, which implies that 
\begin{eqnarray}
\hat{\boldsymbol{\mu}}_1&\xrightarrow{p}&\boldsymbol{\mu}_1, \label{h1}\\
\widehat{\bSigma}&\xrightarrow{p}&\bSigma \nonumber.
\end{eqnarray}
Since $\bSigma$ is a constant matrix, it can be inferred from Slutsky's theorem that
\begin{equation}\label{h2}
\hat{\boldsymbol{\mu}}'_1\widehat{\bSigma}^{-1}\xrightarrow{d}\boldsymbol{\mu}'_1\bSigma^{-1}.
\end{equation}
Notice that $\boldsymbol{x}$ is a constant vector, then
\begin{equation*}
\hat{\boldsymbol{\mu}}'_1\widehat{\bSigma}^{-1}\boldsymbol{x}\xrightarrow{d}\boldsymbol{\mu}'_1\bSigma^{-1}\boldsymbol{x}.
\end{equation*}
From Eq. \eqref{h1} and Eq. \eqref{h2}, applying Slutsky's theorem again, we have
\begin{equation*}
\hat{\boldsymbol{\mu}}'_1\widehat{\bSigma}^{-1}\hat{\boldsymbol{\mu}}_1\xrightarrow{d}\boldsymbol{\mu}'_1\bSigma^{-1}\boldsymbol{\mu}_1.
\end{equation*}
Moreover, both $\boldsymbol{\mu}'_1\bSigma^{-1}\boldsymbol{x}$ and $\boldsymbol{\mu}'_1\bSigma^{-1}\boldsymbol{\mu}_1$ are constants. Therefore,
\begin{eqnarray}     \hat{\boldsymbol{\mu}}'_1\widehat{\bSigma}^{-1}\boldsymbol{x}&\xrightarrow{p}&\boldsymbol{\mu}'_1\bSigma^{-1}\boldsymbol{x},\label{h3} \\
\hat{\boldsymbol{\mu}}'_1\widehat{\bSigma}^{-1}\hat{\boldsymbol{\mu}}_1&\xrightarrow{p}&\boldsymbol{\mu}'_1\bSigma^{-1}\boldsymbol{\mu}_1.\label{h4}
\end{eqnarray}

Next, $r_1$ can be estimated by $\hat{r}_1$, the ratio of number of samples belonging to the first class to the total number of samples. Hence, applying Weak Law of Large Number, we have
\begin{equation}\label{h5}
\ln{\hat{r}_1}\xrightarrow{p}\ln{r_1}.
\end{equation}
Finally, from \eqref{h3}, \eqref{h4} and \eqref{h5}, we conclude that
\begin{equation*}
\hat{\boldsymbol{\mu}}'_1\widehat{\bSigma}^{-1}\boldsymbol{x} - \dfrac{1}{2}\hat{\boldsymbol{\mu}}'_1\widehat{\bSigma}^{-1}\hat{\boldsymbol{\mu}}_1 + \ln{\hat{r}_1}\xrightarrow{p}\boldsymbol{\mu}'_1\bSigma^{-1}\boldsymbol{x} - \dfrac{1}{2}\boldsymbol{\mu}'_1\bSigma^{-1}\boldsymbol{\mu}_1 + \ln{r_1},
\end{equation*}
or
\begin{equation*}
\hat{d}_1\left(\boldsymbol{x}\right)\xrightarrow{p}d_1\left(\boldsymbol{x}\right).
\end{equation*}

\end{document}